\documentclass[letterpaper, 10 pt, conference]{ieeeconf}  

\IEEEoverridecommandlockouts
\usepackage{cite}
\usepackage{amsmath,amssymb,amsfonts}
\usepackage{algorithmic}
\usepackage{graphicx}
\usepackage{textcomp}
\usepackage{xcolor}
\usepackage{url}
\usepackage{comment}
\usepackage{subcaption}
\usepackage{booktabs}
\def\BibTeX{{\rm B\kern-.05em{\sc i\kern-.025em b}\kern-.08em
    T\kern-.1667em\lower.7ex\hbox{E}\kern-.125emX}}
\begin{document}

\title{YCB-Handovers Dataset: Analyzing Object Weight Impact on Human Handovers to Adapt Robotic Handover Motion.
}

\author{Parag Khanna, Karen Jane Dsouza, Chunyu Wang, Mårten Björkman, Christian Smith}
\maketitle

\begin{abstract}
This paper introduces the YCB-Handovers dataset, capturing motion data of 2771 human-human handovers with varying object weights. The dataset aims to bridge a gap in human-robot collaboration research, providing insights into the impact of object weight in human handovers and readiness cues for intuitive robotic motion planning. The underlying dataset for object recognition and tracking is the YCB (Yale-CMU-Berkeley) dataset, which is an established standard dataset used in algorithms for robotic manipulation, including grasping and carrying objects. The YCB-Handovers dataset incorporates human motion patterns in handovers, making it applicable for data-driven, human-inspired models aimed at weight-sensitive motion planning and adaptive robotic behaviors. This dataset covers an extensive range of weights, allowing for a more robust study of handover behavior and weight variation. Some objects also require careful handovers, highlighting contrasts with standard handovers. We also provide a detailed analysis of the object's weight impact on the human reaching motion in these handovers.
\end{abstract}


\section{Introduction}
Handovers are fundamental, yet complex tasks in which humans adapt their motions to different objects based on their properties, such as weight, size, shape and fragility \cite{b1}. Smooth and adaptable handovers that are predictable are essential for promoting natural interactions and improving operational efficiency across various applications, including healthcare, warehouse logistics, and service industries \cite{b2}.

Traditionally, robots have been operated in industrial settings where collaboration was primarily between machines. Recent advancements in robotics call for improved robotic interaction design. Human handover behavior comprises the physical transfer of objects and the interplay of nonverbal cues and gestures that facilitate the exchange. Gaining an understanding of this behavior can offer insights for motion planning in collaborative robots (Cobots). The design of cobots incorporates the use of advanced sensors and safety mechanisms to create a safe collaboration environment while increasing their productivity\cite{b5}. Robots can be taught to adjust their motions in response to varying objects and human cues, ensuring smoother, more natural handovers. Research has shown that incorporating physical cues can increase the safety and effectiveness of handovers in dynamic, real-world applications \cite{b6}.

This paper proposes the novel YCB-Handovers dataset, a motion capture dataset of human-object handovers with varying object weights.  This dataset provides valuable insights into human motion patterns during handovers. 
The supporting dataset for the objects is the well-established YCB Object and Model set. It is commonly used for benchmarking in robotic grasping and manipulation \cite{b3}. It provides a variety of objects with different shapes, textures, and sizes. While significant advancements have been made in human-robot handovers, existing datasets lack the dynamics of handovers involving a wide range of object weights \cite{b4}. Weight adaptation is a critical aspect of developing a collaborative robotic system. Capturing these nuances can help identify critical safety and fluidity properties when applied to human-inspired robotic handover models.
The YCB-Handovers dataset aims to expand on this by capturing human motion during handovers with a wide range of object weights, along with objects with different shapes, sizes and textures, listed in Table \ref{tab:basket_details}.   
\begin{figure}[t]
\centering
\includegraphics[width=0.99\columnwidth,trim={1.3cm 2.5cm 1.3cm 2.6cm},clip]{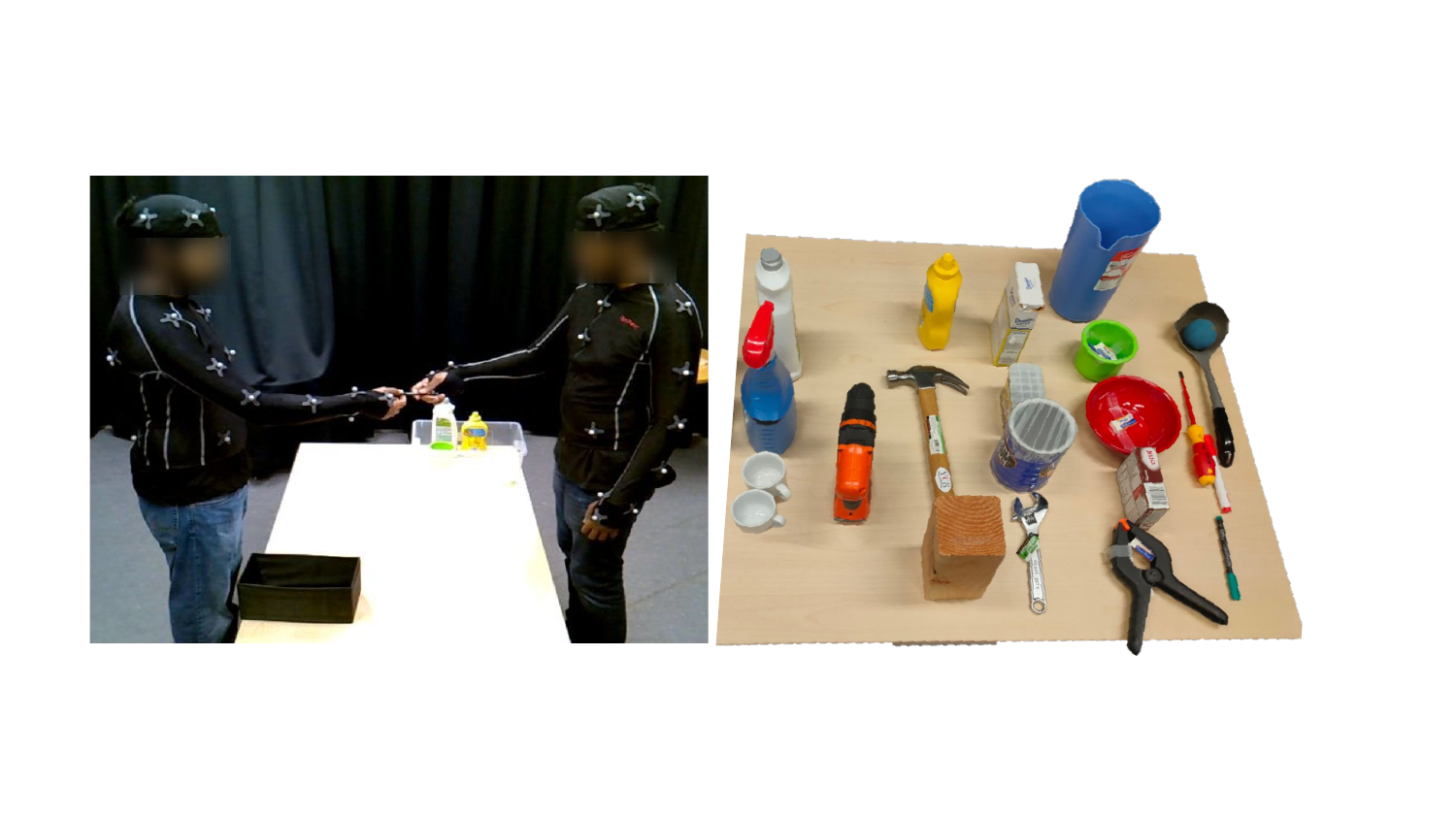}
\caption{YCB-Handovers Dataset: This dataset captures handover interactions involving various objects from the YCB object dataset. The handovers were recorded in a Motion Capture room to ensure accurate tracking of movements. The left image illustrates a handover of a marker pen between two participants, while the right image showcases various objects used in these recorded handovers.}
\label{fig:main_handovers_and_Datasetobjects}
\end{figure}
For this dataset, an experimental study was conducted to record human-human handovers between six pairs of participants comprising different objects. 
Post-processing of the data was conducted to segment 2771 individual handovers from the recorded sequences. Further details of this method are elaborated on in the methodology section of this paper. In summary, this dataset aims to bridge the gap between human handovers with varying weights and real-world human-robot handovers, providing a robust resource for analysing human handovers as well as potentially utilizing data-driven 
models to achieve seamless handovers.

\section{Related Work}
The study of human-robot handovers is critical for developing cobots capable of seamless and intuitive object exchanges with humans. While significant progress has been made in understanding handover dynamics, the specific influence of object weight needs further exploration \cite{survey_review_2022_object_handovers}. This section reviews existing datasets, methodologies, and adaptive strategies, emphasizing the necessity of weight-sensitive data for robotic handover studies.

Chao et al. presented HandoverSim \cite{b7}, a Python-based simulation environment and benchmark for human-to-robot object handovers. This framework facilitated the study of handover mechanics in a controlled setting, allowing for the evaluation of various handover strategies. However, it primarily focuses on simulation and may not fully capture the complexities of real-world, weight-sensitive handovers. The Object Hand-Over (OHO) dataset \cite{b8}, comprising multi-modal data such as color, depth, and thermal images of various objects held by human hands, was introduced to support machine learning tasks such as object recognition.

The H2O dataset presented by Ye et al. consists of annotated videos of handovers between 15 individuals for over 30 different objects \cite{b9}. This dataset supports vision-based tasks and provides insight into human-human handover dynamics. But, it lacked insights based on weight diversity. Similarly, the Human-Object-Human (HOH) dataset, encompassing 2,720 handovers involving 136 objects and 40 participants, provides multi-modal data to analyze handover mechanics but does not explicitly include object weights \cite{b10}.

Recent studies also show various methods to improve human-robot handovers. Aleotti et al. proposed a system that considers object affordances to optimize the orientation and positioning of objects during robot-to-human handovers \cite{b11}. This approach was meant to facilitate handovers such that the most convenient part of the object was faced towards the taker. Chang et al in \cite{b12} presented a deep reinforcement learning-based method for learning 6 degrees of freedom grasp choices in human-to-robot handovers. Their system enables robots to predict human-preferred grasps, enhancing the naturalness of handovers. Finally, other methods presented by Lori et al. use Dynamic Movement Primitives and Preference Learning to adapt control in human-robot handovers \cite{b13,b14}.

\section{Methods}
The methods used within this study can be broken down into several phases to give an insight into the design, data collection, and post-processing procedures in creating the YCB-Handovers dataset.

\subsection{Experimental Design}
The dataset was generated by recording handovers between six participant pairs, i,e, a total of twelve participants.
We had one right handed-left handed pair while all other pairs were right handed-right handed pair. Each pair conducted handovers for five different group of objects, presented in a basket. 
Each of the four baskets holds a variety of objects so that the movements and handovers would be representative of varying weight classes. The specific details of the baskets are mentioned within Table \ref{tab:basket_details}. 
During the final round (the fifth basket), we introduced four unique objects to observe both careful and non-careful human-to-human handovers, along with the impact of added weights. Previous research 
showed that carefulness in handover motions can be identified in real-time when transferring objects like filled versus empty cups \cite{careful_handovers_Lastrico}. Building on this, we used a measuring cup (weighing 0.008 kg and 0.048 kg when filled with water) and the YCB pitcher, filled either with water or additional weights. This setup enabled us to study how a significant increase in object weight (2 kg in this case) affects human handover motions involving objects that require careful handling. 
\begin{figure}[h]
\centering
\subfloat[Basket 1]{\includegraphics[width=0.42\linewidth]{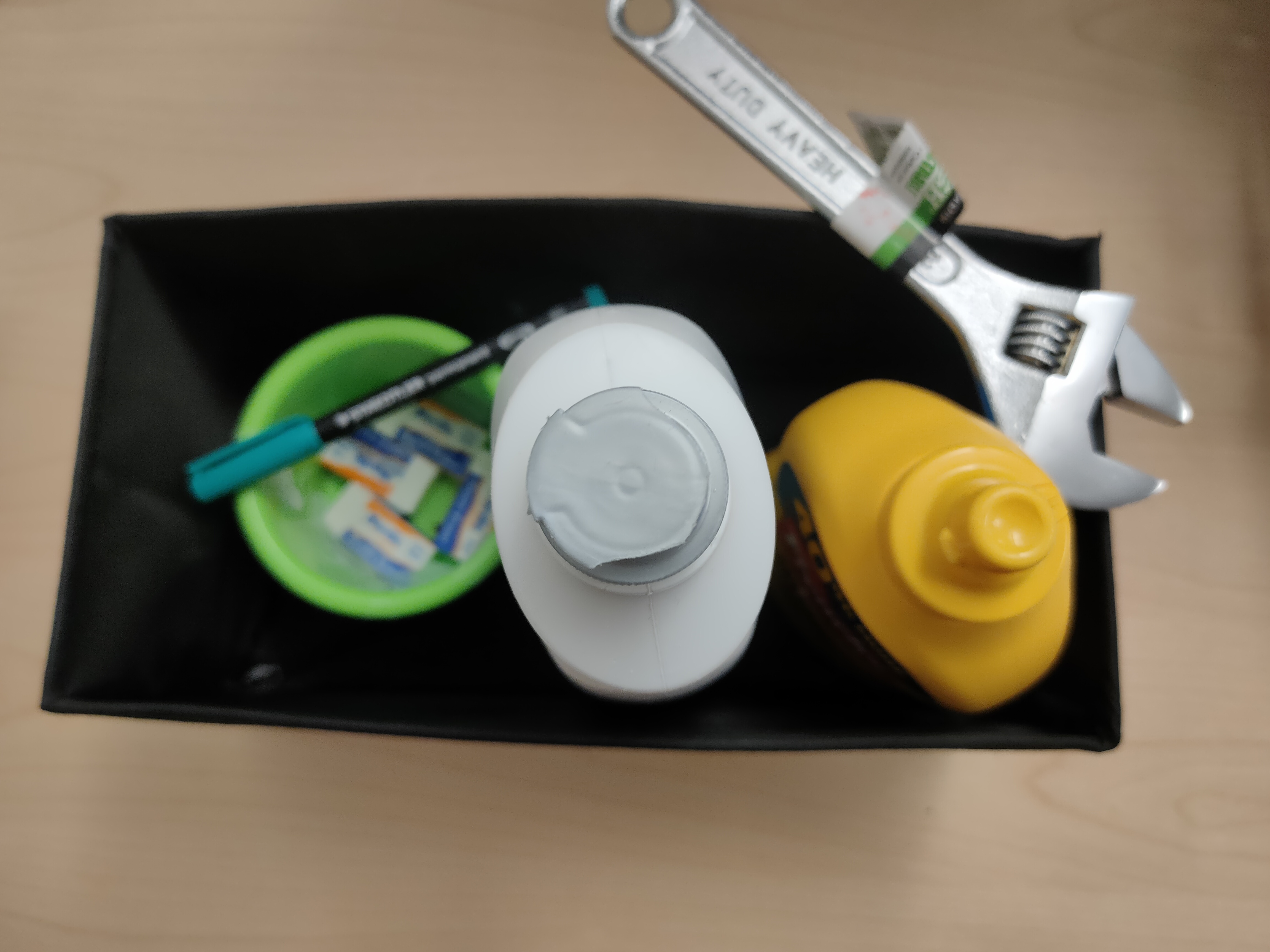}\label{fig:subfig1}}
\hspace{2mm}
\subfloat[Basket 2]{\includegraphics[width=0.42\linewidth]{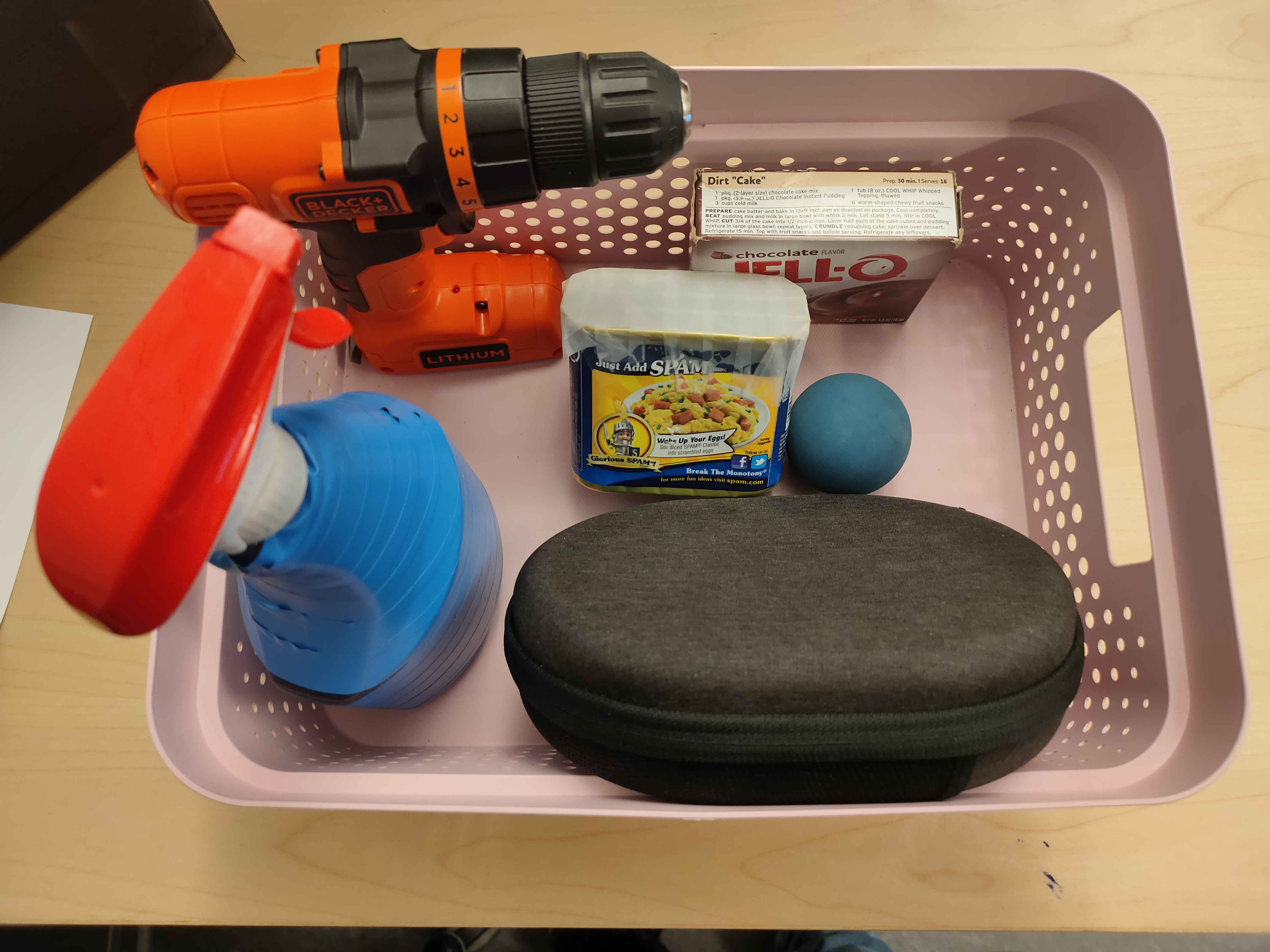}\label{fig:subfig2}}
\hfill
\subfloat[Basket 3]{\includegraphics[width=0.42\linewidth]{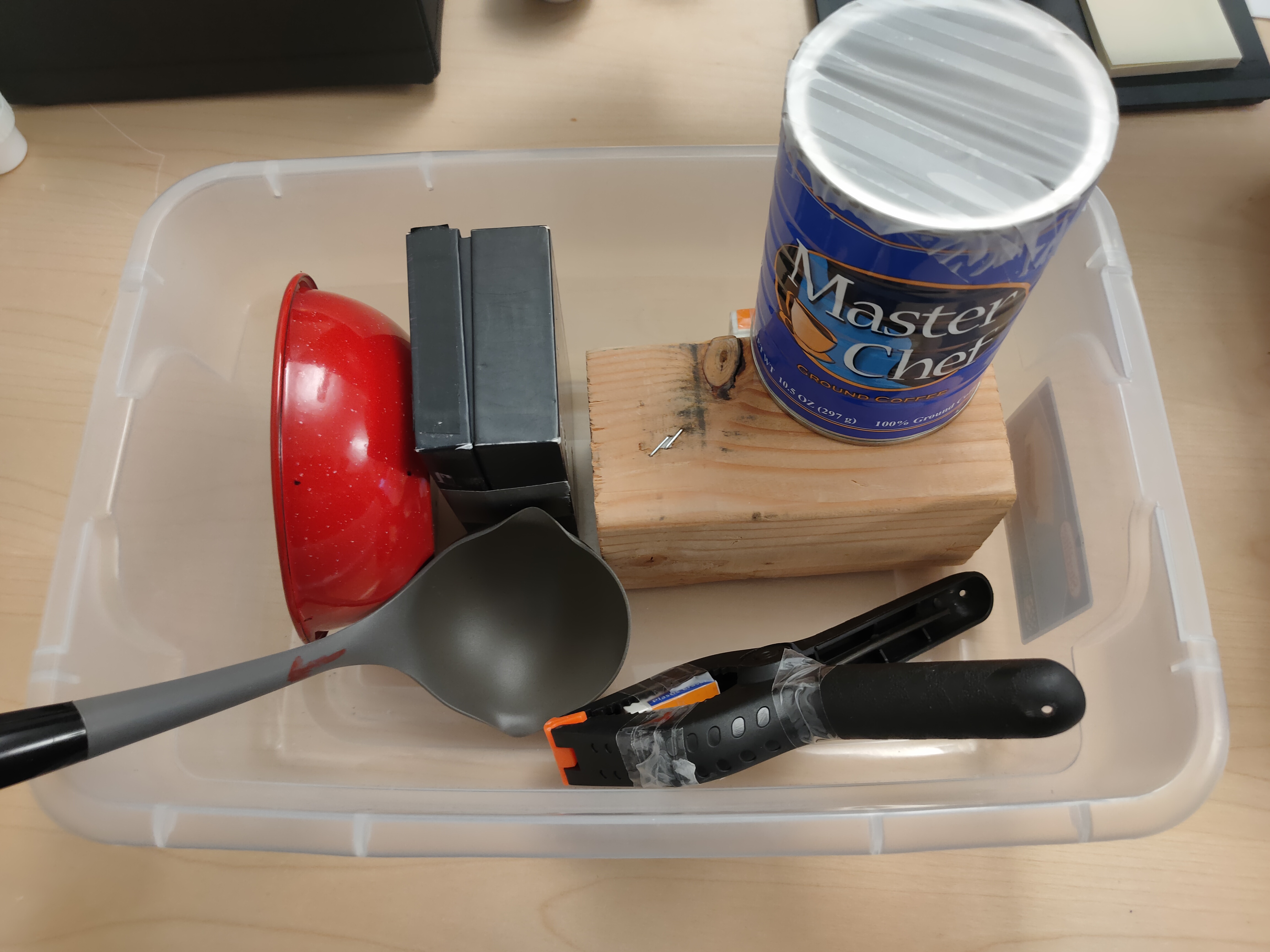}\label{fig:subfig3}}
\hspace{2mm}
\subfloat[Basket 4]{\includegraphics[width=0.42\linewidth]{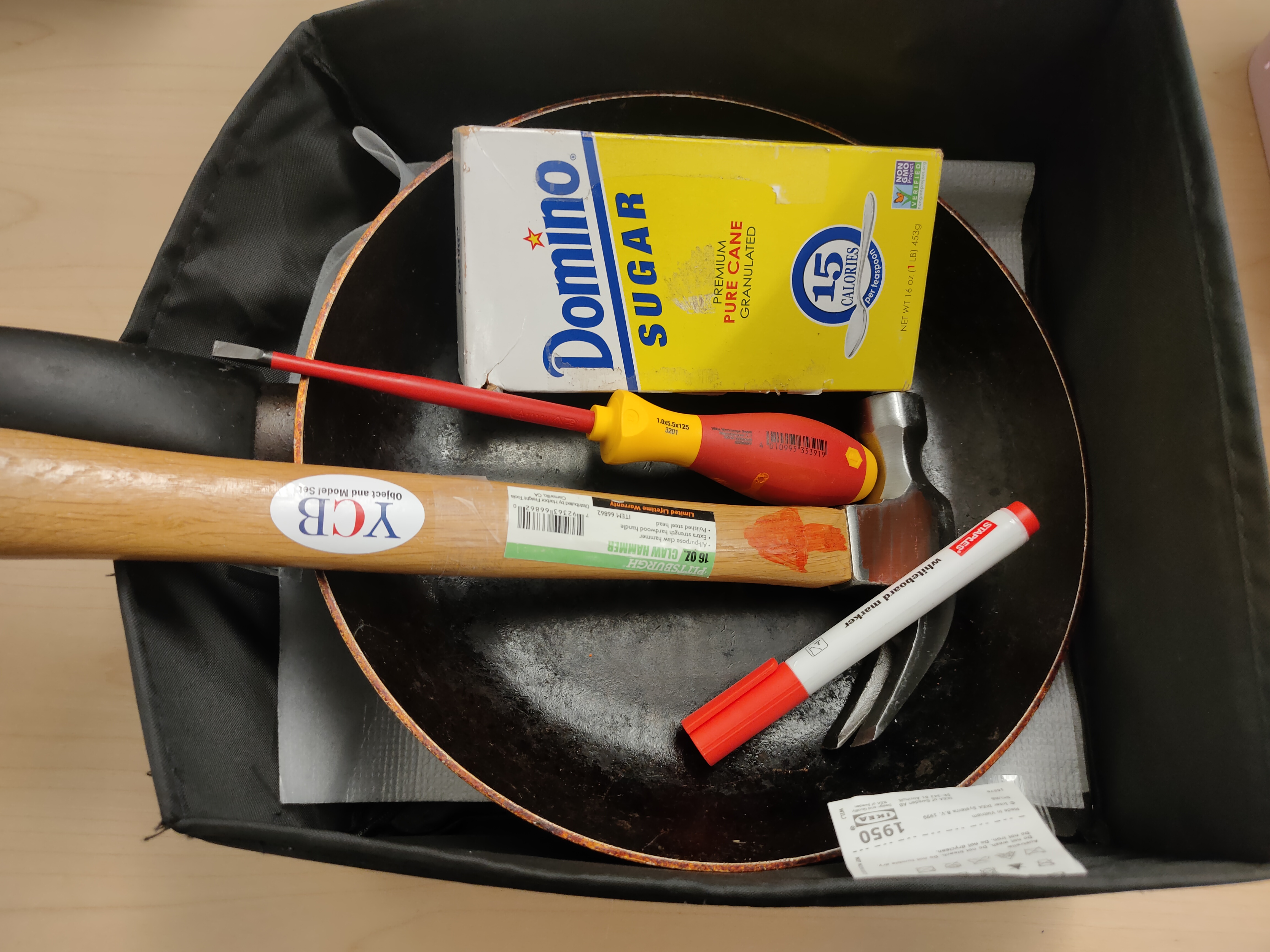}\label{fig:subfig4}}
\caption{Different objects used in the the different baskets}
\label{fig1}
\end{figure}

The experiment was conducted in five rounds, with each round corresponding to one basket. 
The setup involved a pair of participants performing handovers in a controlled environment, where their movement was captured using in a motion capture (MoCap) room using MoCap suits, utlizing a MoCap system by Optitrack.
This captured the spatial trajectories of markers placed on the participants' bodies. Each participant was assigned the role of either `Skeleton 1' or `Skeleton 2' to maintain consistency and differentiability in the data. This helps conduct data validation by identifying the participant who started the sequence of a series of handovers when conducting segmentation described later in this section.
Two participants stood facing each other across a table as shown in Fig. 2. In each round, participant 1 had a basket filled with objects placed near their dominant hand, while participant 2 had an empty basket positioned near their dominant hand. Participant 1 would transfer the objects one at a time from their basket to participant 2, who then placed them in their own basket. Once participant 1 finished handing over all the objects, participant 2 would start transferring objects from their basket back to participant 1. This cycle continued for 9 minutes per basket.
For the final round (fifth basket), the participants were given the objects one by one and they performed the repeated handovers for 2 minutes per object.

To ensure natural handover behaviour, the participants were instructed to handle objects as they normally would using their preferred or dominant hand. They were encouraged to talk during the handovers to maintain the naturalness of the handovers, as done in another similar study \cite{b15}. 
Four of these baskets are shown in Fig. \ref{fig1}.

\begin{table*}[ht]
\centering
\setlength\abovecaptionskip{-0.2\baselineskip}
\caption{YCB Handovers: Details of Objects in Different Baskets}
\begin{tabular}{|c|c|c|c|}
\hline
\textbf{Basket} & \textbf{No. of Handovers} & \textbf{Weight (kg)} & \textbf{Object Details} \\ \hline
1 & 128 & 0.008 & Marker Small (YCB) \\ 
  & 128 & 0.118 & Mug (YCB) \\ 
  & 128 & 0.242 & Wrench (YCB) \\ 
  & 126 & 0.600 & Mustard Bottle (YCB) \\ 
  & 128 & 1.131 & Cleanser Bottle (YCB) \\ \hline
2 & 108 & 0.040 & Racquetball (YCB) \\ 
  & 106 & 0.190 & Jello-Choc Box (YCB) \\ 
  & 106 & 0.374 & Meat Can (YCB) \\ 
  & 107 & 0.874 & Hand Drill (YCB) \\ 
  & 107 & 1.020 & Spray Bottle (YCB) \\ 
  & 103 & 1.450 & Earphone Cover with Weights  \\ \hline
3 & 95 & 0.055 & Spatula (YCB) \\ 
  & 101 & 0.149 & Bowl (YCB) \\ 
  & 99  & 0.202 & Clamp (YCB) \\ 
  & 101 & 0.410 & Coffee Can (YCB) \\ 
  & 100 & 0.728 & Wood Block (YCB) \\ 
  & 100 & 1.300 & Black Box with Weights \\ \hline
4 & 99 & 0.015 & Large Marker (YCB) \\ 
  & 98 & 0.095 & Screwdriver (YCB) \\ 
  & 100 & 0.183 & Pitcher Base (YCB) \\ 
  & 101 & 0.514 & Sugar Box (YCB) \\ 
  & 100 & 0.608 & Hammer (YCB) \\ 
  & 85  & 0.925 & Skillet (YCB) \\ \hline
5 & 91 & 2.060 & YCB Pitcher with Added Weights (Heavy Weight-Not Careful) \\ 
  & 69 & 2.060 & YCB Pitcher with Water (Heavy Weight-Careful) \\ 
  & 96 & 0.008 & Measuring Cup (Light Weight-Not Careful) \\ 
  & 61 & 0.048 & Measuring Cup with Water (Light Weight-Careful) \\ \hline
\end{tabular}
\label{tab:basket_details}
\end{table*}

The participants were instructed to wear motion capture suits with reflective markers placed at the key joints such as the wrists, elbows, shoulders, and the head and chest. Only the data from the upper torso is captured to focus on the handover dynamics. 
This method of recording allowed for the handover process to be captured in 3D space along with a time index. This allowed the calculation of the associated positions and velocities of the joints.

\subsection{Data Collection}
At the start of each handover session, the cameras within the MoCap room were calibrated using the calibration wand for spatial alignment. This ensured accurate tracking of the positions of the joints of both the participants. The calibration also mitigated the potential drift or misalignment over time \cite{b15}. A trial signifies one series of handovers. Before each trial is started, one of the baskets is placed on the table at the centre of the room. The participants are then instructed to stand on either side of the table and their positions are verified with the cameras. During the trial, the participants are encouraged to be as natural as possible when conducting the handovers to avoid any biases and errors within the extracted data.

The motion capture system collect spatial trajectories of all the markers over time with their unique labels. Within the Optitrack recording software, a skeleton is formed and the markers are associated with the respective parts of the skeleton. This way, every reflective marker is tracked as part of the whole body.
The rosbag format was used to enable the capture of synchronised data streams from the cameras. This format was selected due to the storage of properties such as image data, time-stamped markers, and object metadata as rostopics that can be referenced during data extraction and processing. 
This format facilitated efficient data storage and time-synchronisation. All the visual, positional and temporal data streams are synchronised across the trials to provide a reliable time index. This makes the dataset ideal for use in time series analysis. An extract from the handover data can be seen in Fig. 3.

Finally, a Python script was run to extract the rosbags into \textit{.csv} files for each of the body segements in the two skeletons for the participant pair. 
Additionally, these files also divided the giver and the taker as separate `skeletons' to allow for the identification of the data. 
The extracted files were then verified and prepared for post-processing to be done in MATLAB.

\subsection{Data Post-Processing}

The next phase after data collection involves segmentation of the data into discrete handovers while extracting features. The main focus of this section is to extract meaningful segments. The data collected from the motion capture system is in the form of raw marker trajectories. The cameras provided precise positional data in three-dimensional space for each marker. They also provided time steps, allowing for velocity analysis of the joints. A custom MATLAB script was used to manage data loading, synchronisation, segmentation and feature extraction for both the participants based on proximity and time thresholds. 


The main parts of interest during the segmentation are the dominant hands of the participants. The pose data of the dominant hands is first separated out.
The pose data is visualised in Fig. 3 and is used to find the relationship between the dominant hands of the two participants.
When a handover is being been performed, the dominant hands of the giver and the taker come close in proximity when both the giver and the taker are holding the object. Using this proximity, a threshold can be used to segment the handovers. In simple words, the handovers are segmented at the point when both participants are holding the object.

\begin{figure}[h]
\centering
\includegraphics[width=0.9\columnwidth,trim={2.6cm 8.5cm 2.6cm 2.6cm},clip]{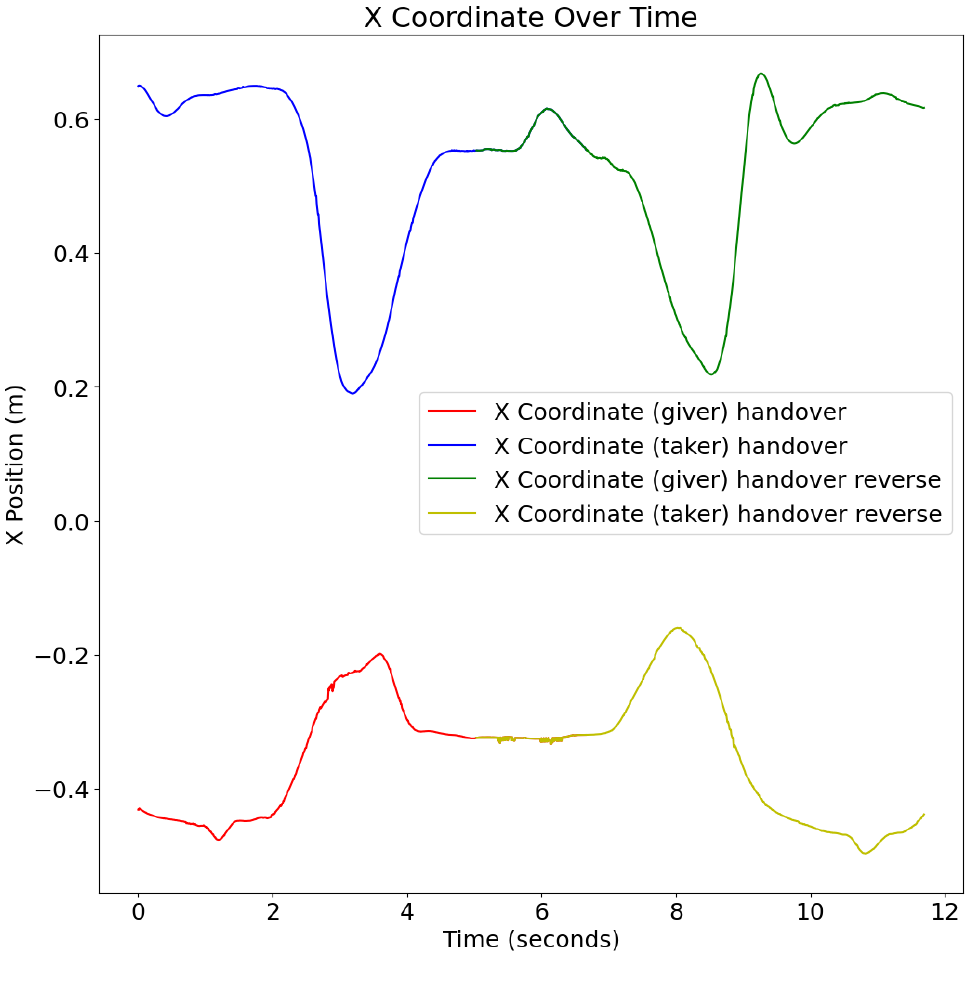}
\caption{The image shows the position of the dominant hand during two consecutive handovers of Pair 1 for the non-careful trial with the YCB pitcher.}
\label{fig3}
\end{figure}
Finally, for each detected handover, the script extracts the spatial positions for the upper torsos of the giver and the taker and segments the data into separate files, saving them as .csv files. 
These files also contain extracted images to verify the point of handover and to identify the object, however, are not made public with the dataset due to anonymization requirements.
These images were used for robust manual data validation used for labeling the handovers with the object labels. This post-processing pipeline aligns with methodologies in similar studies which concern the capture of handovers \cite{b15}. This structured approach enhances the YCB-Handovers dataset's utility in training machine learning models. Thus, Fig. 3 shows extracted consecutive handovers, showing the positional trajectory of the participants' dominant hand.

\section{Dataset Description} \label{Dataset Description}
The YCB-Handovers dataset consists of motion capture recordings from object handovers. This dataset includes a variety of objects with varying weights, sizes and shapes. The file formats stored within this dataset include \texttt{.csv} files for each body segment of the tracked upper body skeleton of the participants.
The typical segments stored within one of the extracted \texttt{.csv} files can be seen in Table \ref{tab:dataset_description}. These features are stored for both the giver and the taker.
This dataset shows the potential to be used with time series analysis methods for the training of robotic handover models. Each handover in the dataset is labelled according to the object used in handover, to facilitate learning of such human inspired handover models.

\begin{table}[h!]
\setlength\abovecaptionskip{-0.2\baselineskip}
  \caption{Segments Recorded in a Skeletal Representation for the Giver and the Taker in a Handover}
  \label{tab:skeleton1}
  \resizebox{0.99\columnwidth}{!}{
  \begin{tabular}{cccc}
    \toprule
    
    
    Segment & Segment & Segment & Segment \\   
     No. & Name & Frame & Components\\
    
    \midrule
    1 & hip\_pose & Hip & Position(x,y,z)\\
     & & & {Rotation}$^*$($q_0,q_1,q_2,q_3$)\\
    2 & ab\_pose & Ab & ---\texttt{"}---\\
    3 & chest\_pose & Chest & ---\texttt{"}---\\
    4 & neck\_pose & Neck & ---\texttt{"}---\\
    5 & head\_pose & Head &  ---\texttt{"}---\\
    6 & LShoulder\_pose & Left Shoulder &  ---\texttt{"}---\\
    7 & LUArm\_pose & Left Upper Arm &  ---\texttt{"}---\\
    8 & LFArm\_pose & Left Forearm &  ---\texttt{"}---\\
    9 & LHand\_pose & Left Hand &  ---\texttt{"}---\\
    10 & RShoulder\_pose& Right Shoulder &  ---\texttt{"}---\\
    11 & RUArm\_pose & Right Upper Arm &  ---\texttt{"}---\\
    12 & RFArm\_pose & Right Forearm &  ---\texttt{"}---\\
    13 & RHand\_pose & Right Hand &  ---\texttt{"}---\\
    \bottomrule
    \multicolumn{4}{c}{*Each pose message contains 3 position and 4 quaternion rotation components} \\
    \end{tabular}
    \label{tab:dataset_description}
}
\end{table}

With regard to the limitations of the generated dataset, errors might be noticed where certain body markers are temporarily obscured due to the hand movements and the participant's positioning. Such handovers were removed from the dataset. Noise might also be present in the positional data. This can be addressed by applying suitable filtering techniques. 
Finally, during the handovers, unintended arm movements may have been recorded during the approach and retreat phases in a handover, leading to inconsistencies when training a model.

\textbf{Usage Notes}:
The dataset is intended for researchers investigating human-human handovers to inspire human-robot handovers. 
Participants were enlisted through advertisements displayed on the university campus. Each participant was rewarded with a 100 SEK gift voucher for taking part.
As per the local regulations, no ethical approval was required for this study, as no sensitive personal data (such as racial/ethnic origin, political views, religious/philosophical beliefs, health/sexual life) was collected, and the research did not involve any physical interventions or biological samples from participants. In the absence of a relevant ethics committee, the study adhered to the principles outlined in the Declaration of Helsinki.
Participants initially completed a consent form for data collection and reviewed the study instructions. They explicitly agreed to the use and dissemination of their anonymized data, as well as the use of recorded video data for academic publications and presentations.
The dataset is available at: \url{https://github.com/paragkhanna1/YCB-Handovers}. 

\section{Analysis: Impact of Object Weight on motion}
To make the motion capture data less noisy, we used a 4th-order Butterworth filter with a 5 Hz cutoff frequency. This helped to eliminate high-frequency noise caused by inaccuracies in the motion capture system or tracking issues. This analysis excluded objects requiring careful handovers.

\subsection{Velocities and Acceleration}

\begin{figure}[h!]
      \centering
        \includegraphics[width=0.9\columnwidth,trim={0.5cm 0.5cm 0.5cm 0.3cm},clip]{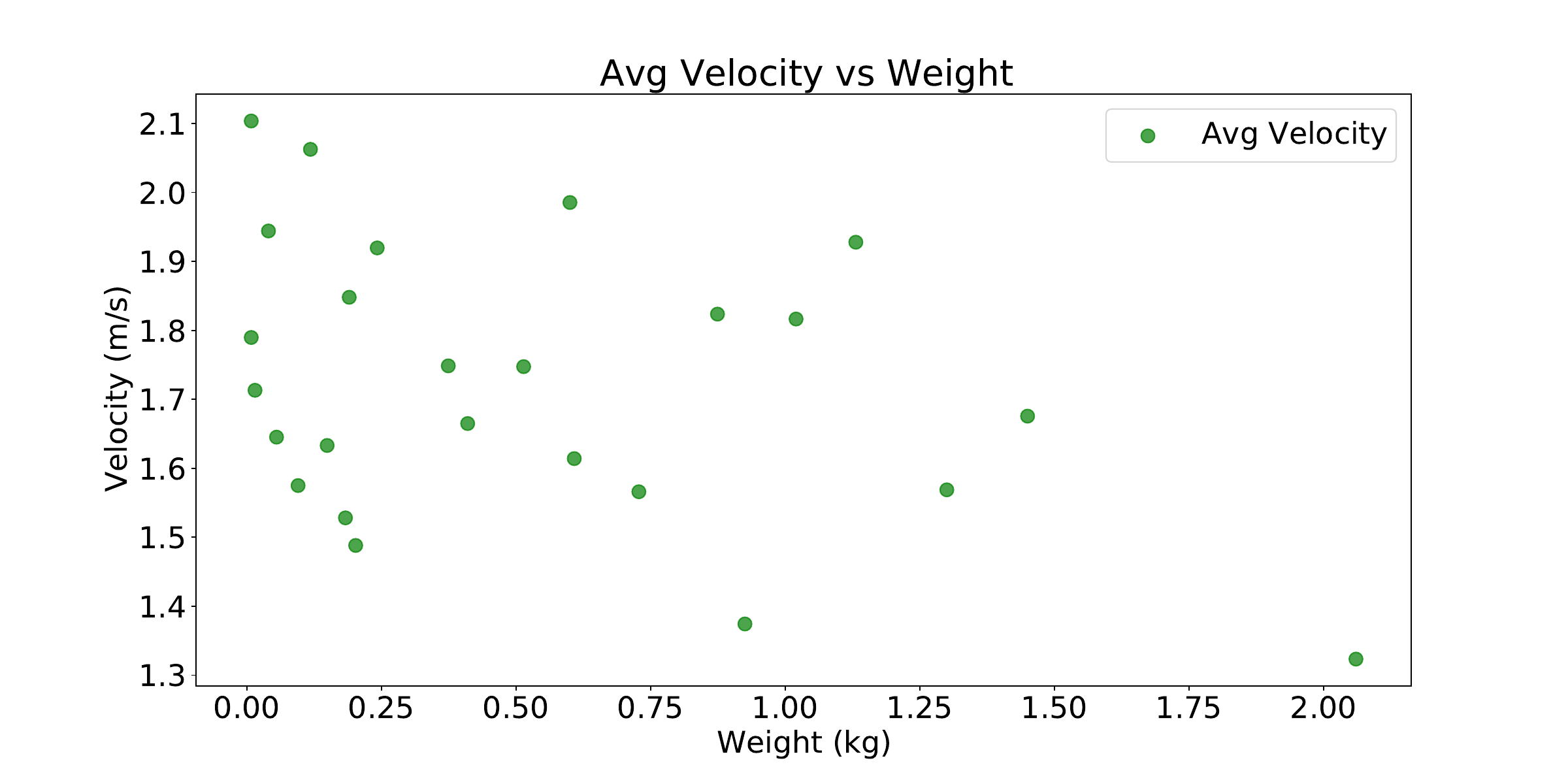}
        \vspace{0.05cm}
        \includegraphics[width=0.9\columnwidth,trim={0.5cm 0.5cm 0.5cm 0.3cm},clip]{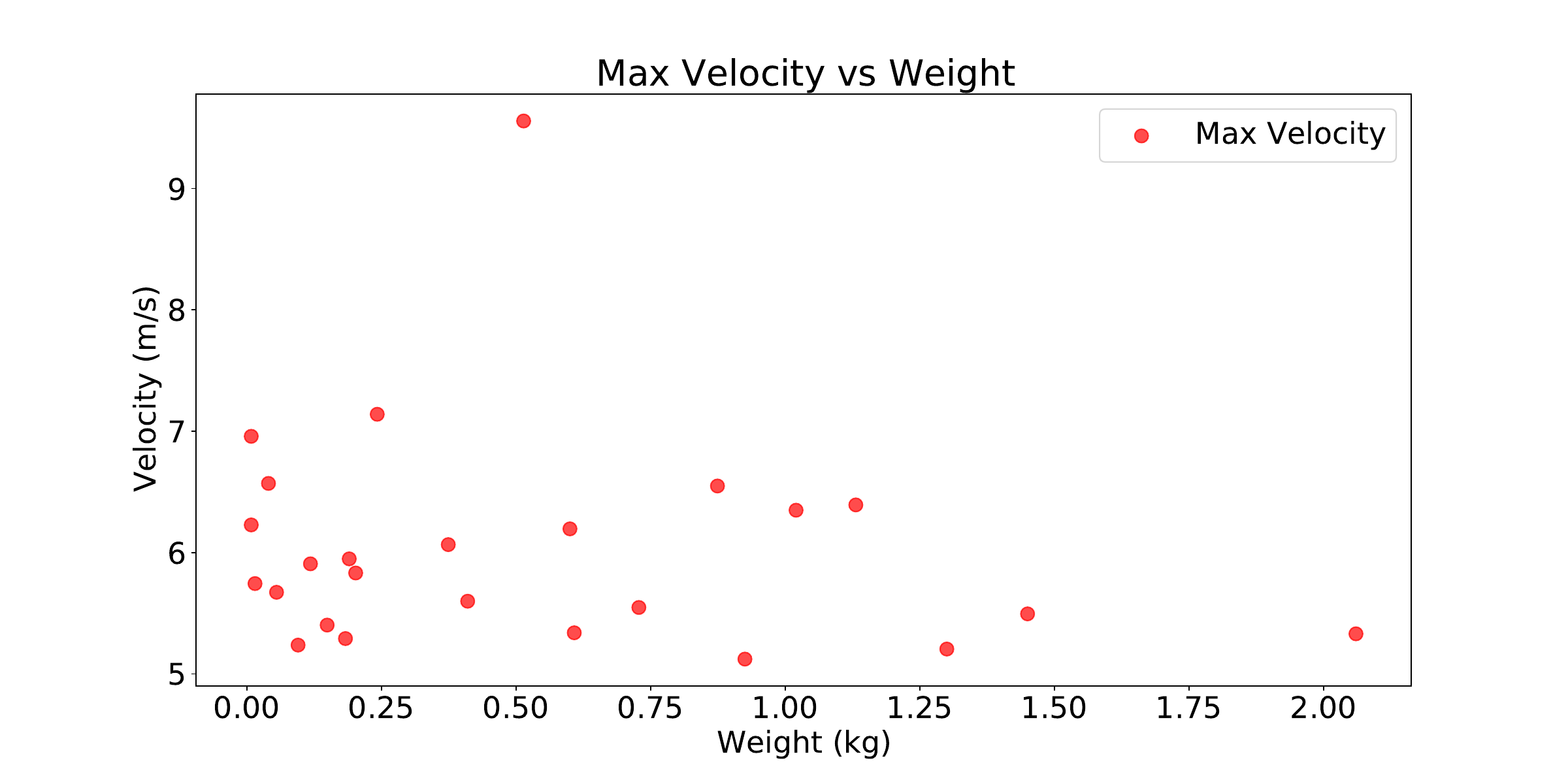}

        \includegraphics[width=0.90\columnwidth,trim={0.5cm 0.5cm 0.5cm 0.3cm},clip]{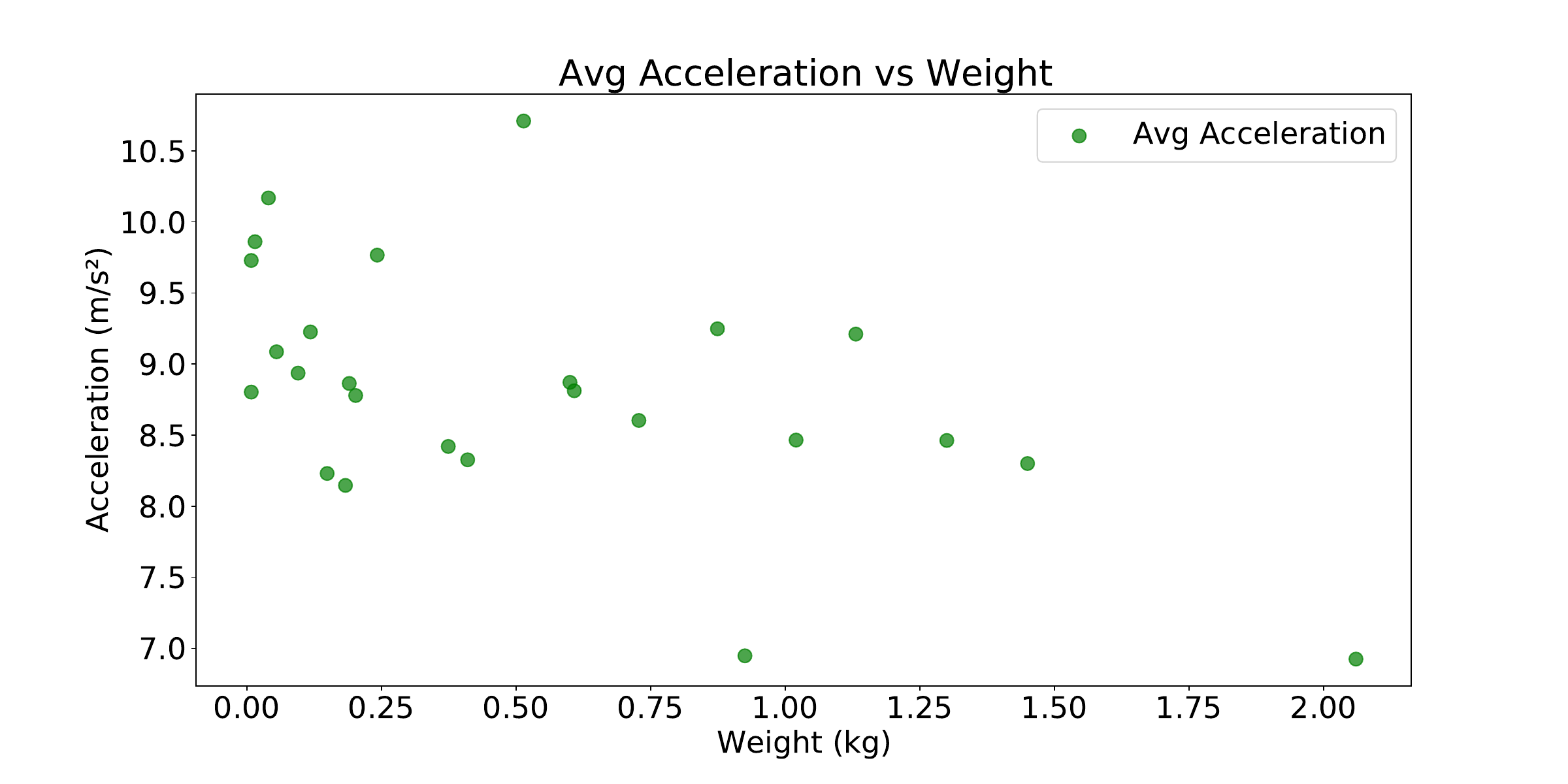}

        \includegraphics[width=0.90\columnwidth,trim={0.5cm 0.5cm 0.5cm 0.3cm},clip]{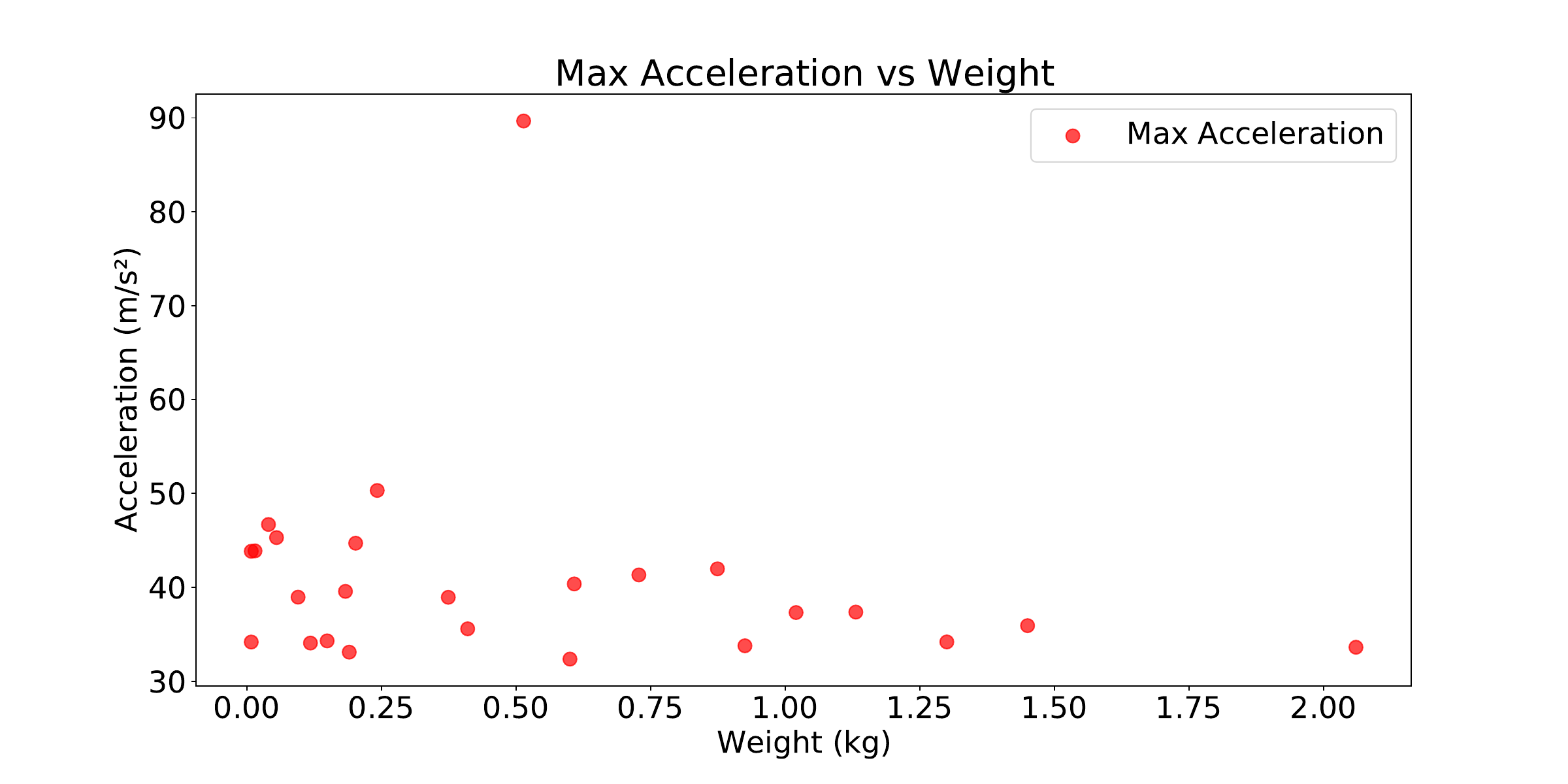}
    \setlength\abovecaptionskip{-0.01\baselineskip}
      \caption{Motion characteristics for human hand motion with different weighted objects in human-human handovers. Mean values observed for a particular object are plotted. Note: This plot excluddes the objects with added carefullness}
      \label{figC:motion_avg_max_vel}
   \end{figure}
\begin{figure*}[h]
      \centering
        \includegraphics[width=0.9\columnwidth,trim={0cm 0.0cm 0cm 0.3cm},clip]{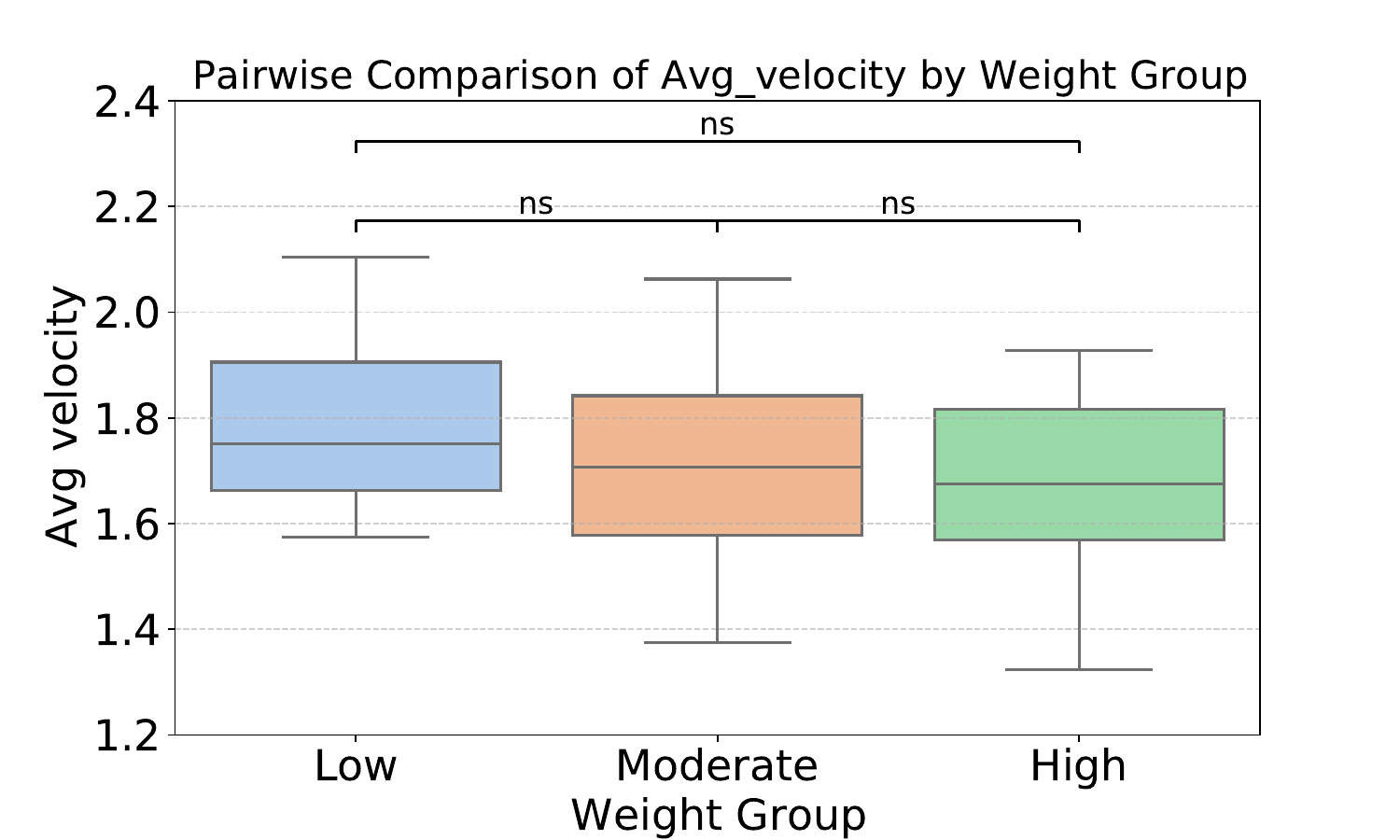}
        \includegraphics[width=0.9\columnwidth,trim={0cm 0.0cm 0.3cm 0.3cm},clip]{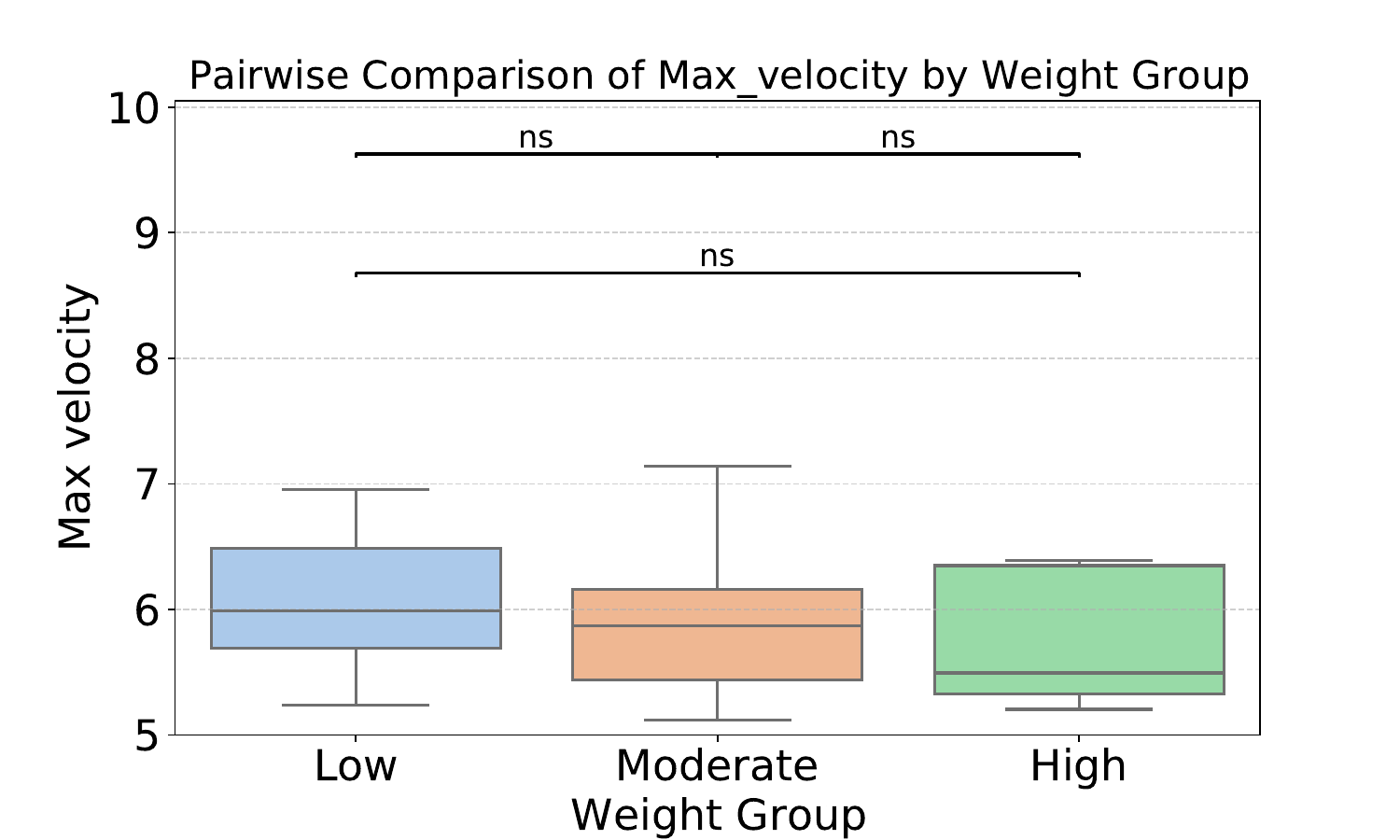}
        \vspace{0.5cm}
        \includegraphics[width=0.9\columnwidth,trim={0cm 0.0cm 0.3cm 0.3cm},clip]{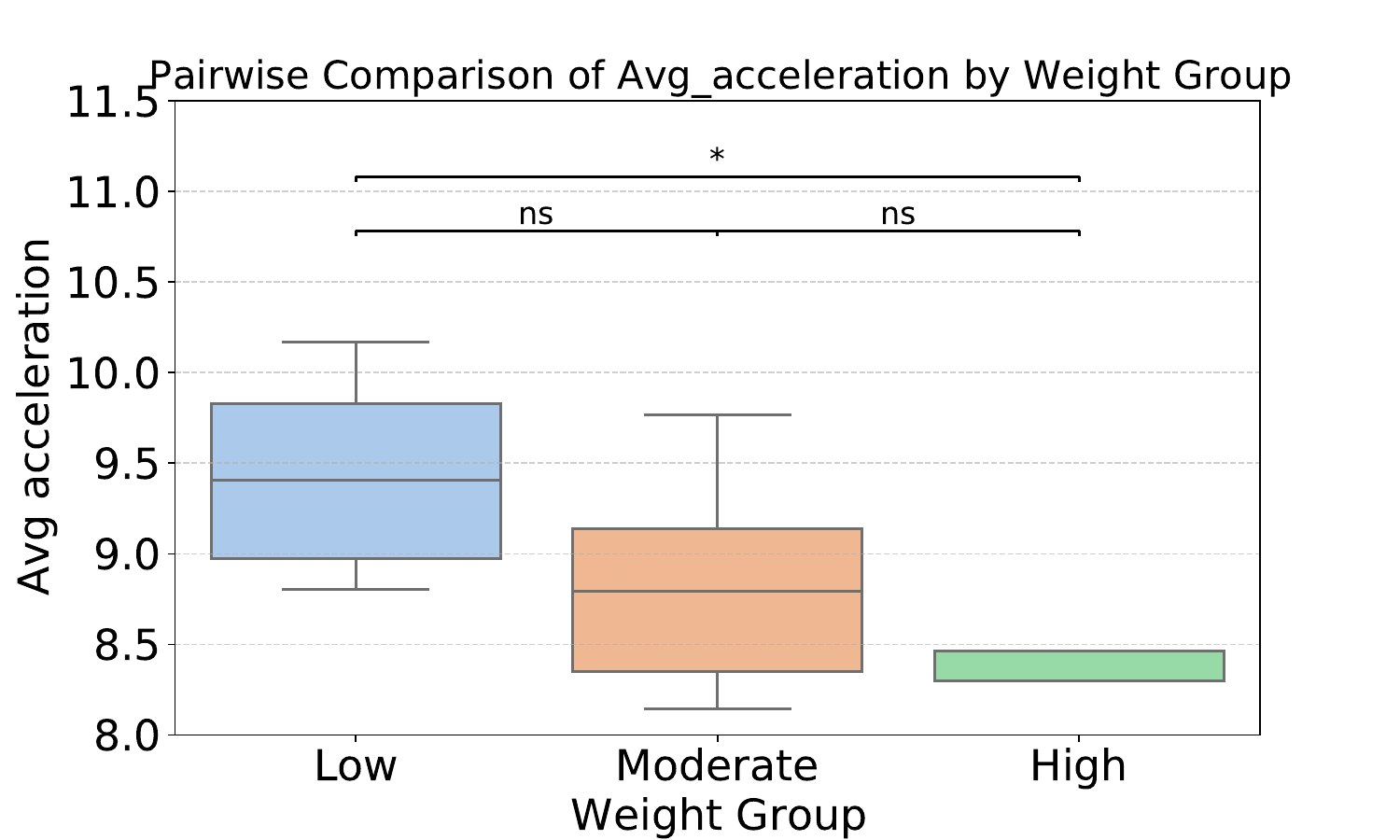}
        \includegraphics[width=0.9\columnwidth,height=4.80cm,trim={0cm 0.0cm 0.1cm 0.3cm},clip]{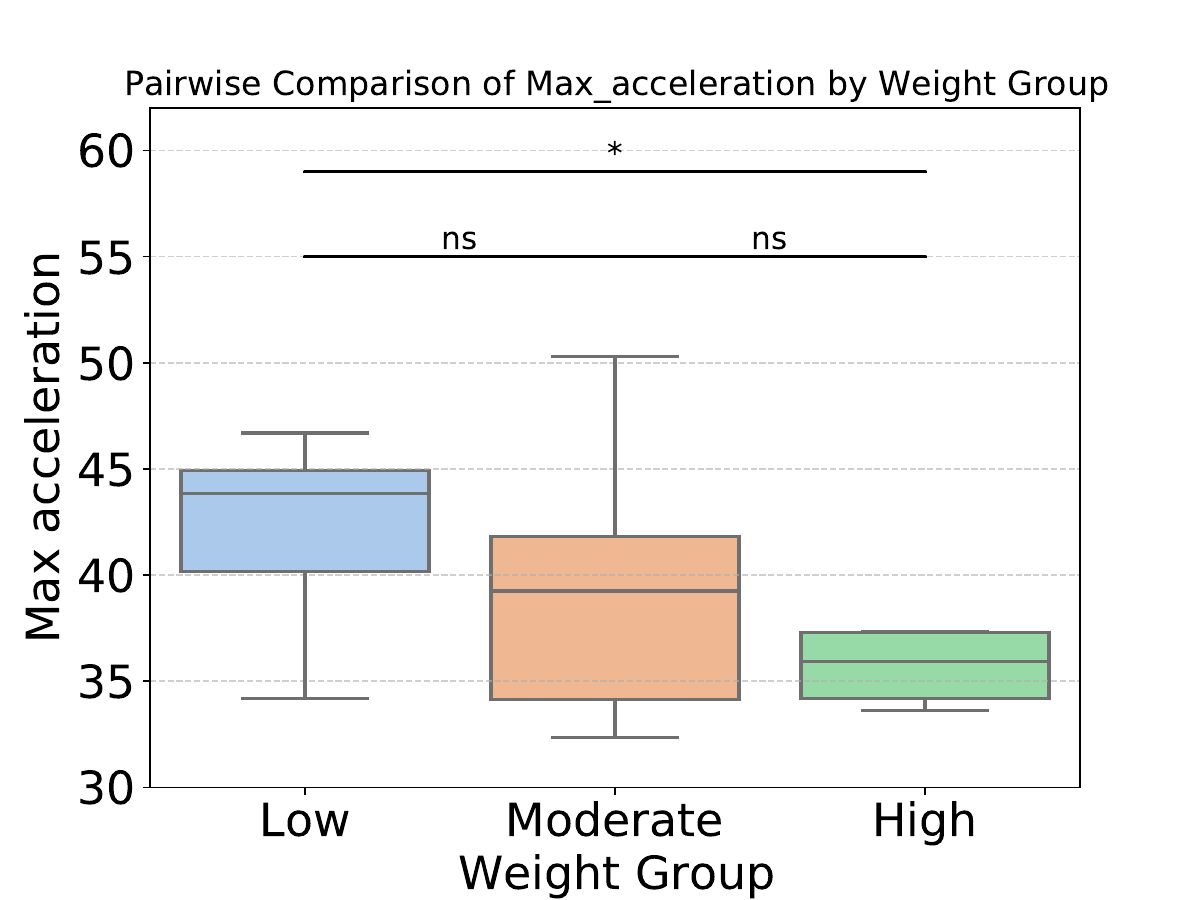}
    \setlength\abovecaptionskip{-1.1\baselineskip}
      \caption{Motion Characteristics across different Weight Categories}
      \label{fig:motion_box_plots_3categ}
   \end{figure*}

We studied the motion during the ``reaching phase'' when human givers moved their hands after picking up an object to transfer it. 
Specifically, we looked at the hand's position in 3D (x, y, z coordinates), as well as the average and maximum velocitiees and accelerations of the hand during the handover for each object. These details are presented in Fig. 4, excluding objects with added carefulness. Generally, we observed that heavier objects led to slower movements and lower acceleration. However, there were some exceptions, where factors like object type, shape, and size also affected the giver's hand motion.
  
\textbf{Correlation with Object Weight}:

Analyzing the relationship between object weight and motion characteristics (velocity and acceleration) for hand motion reveals a negative correlation between motion characteristics and weight.
\begin{table}[h!] 
\centering
\setlength\abovecaptionskip{-0.07\baselineskip}
\caption{Correlation with Weight for Various Metrics}
\begin{tabular}{lcc} \hline & \textbf{weight} \\ \hline \textbf{avg\_velocity} & -0.393 \\ \textbf{max\_velocity} & -0.170 \\ \textbf{avg\_acceleration} & -0.538 \\ \textbf{max\_acceleration} & -0.184 \\ \hline 
\end{tabular} 
 \label{tab:correlation_with_weight} \end{table}

The relationships between object weight and various metrics are shown in the Table \ref{tab:correlation_with_weight}. A correlation coefficient (\(r\)) is classified as strong if \(|r| \geq 0.7\), moderate if \(0.5 \leq |r| < 0.7\), weak if \(0.3 \leq |r| < 0.5\), and very weak or negligible if \(|r| < 0.3\).
According to these classifications, the average acceleration (\(r = -0.538\)) has a moderate negative correlation with weight, meaning that average acceleration tends to decrease as weight increases. The correlations between weight and average velocity, maximum velocity, and maximum acceleration are all weak negative correlations. This suggests a slight tendency for these metrics to decrease with heavier objects.

In summary, average acceleration shows the strongest link with weight, but none of the metrics have a strong correlation. This indicates that the relationship between weight and motion characteristics varies across individuals. This highlights how complex human handovers are and how factors like individual differences might influence hand motion when transferring objects of different weights.

\subsection{Weight Groups}
For further analysis, all objects in Table~\ref{tab:basket_details} were categorized into three weight groups: Low, Moderate, and High. 

Different weight limits for these categories were explored to identify clear and significant differences in motion characteristics, particularly velocities and accelerations. This method helped distinguish the three weight categories effectively, forming a solid basis for further analysis. The box-plots in Fig.~\ref{fig:motion_box_plots_3categ} compare motion metrics for reaching motions during human handovers across the three object categories, revealing significant differences in some metrics.

For Low vs. High weight categories, significant differences were observed in average acceleration (\(p = 0.002\)) and maximum acceleration (\(p = 0.004\)). These findings highlight consistent differences in velocity and acceleration for human handover motions across objects of increasing weights.

\begin{table}[h!]
\centering
\setlength\abovecaptionskip{-0.07\baselineskip}
\caption{Average and Maximum Velocities and Accelerations by Weight Group}
\resizebox{0.995\columnwidth}{!}{
\begin{tabular}{lcccc}
\hline
\textbf{weight\_group} & \textbf{avg\_vel} & \textbf{max\_vel} & \textbf{avg\_acc} & \textbf{max\_acc} \\
\hline
Low & 1.79 & 6.06 & 9.43 & 42.14 \\
Moderate & 1.71 & 6.10 & 8.78 & 42.14 \\
High & 1.66 & 5.75 & 8.27 & 35.68 \\
\hline
\textbf{Avg weight} & 0.03 & 0.43 & 1.39 \\
\hline
\end{tabular}
}
\label{tabT:vel_acc_weights}
\end{table}
Considering velocity trends, low-weight objects (0.008–0.1 kg) have the highest velocities (avg: 1.79 m/s, max: 6.06 m/s), while high-weight objects (0.950–2.060 kg) have the lowest (avg: 1.66 m/s, max: 5.75 m/s). Moderate-weight objects fall in between.
For acceleration trends, low-weight objects exhibit the highest accelerations (avg: 9.43 m/s², max: 42.14 m/s²), while high-weight objects have the lowest (avg: 8.27 m/s², max: 35.68 m/s²). Moderate-weight objects again lie in between.
Overall, velocity and acceleration decrease with increasing weight, with the most significant drop in the high-weight category.

\subsection{Clustering for Weight Classification}


Clustering is a basic method in machine learning that helps us group similar items together. Using clustering, we do weight classification to understand how different weights affect human handovers. This can help us see patterns in how humans move or respond when handling objects of different weights. By organizing objects into weight groups, we can improve handover strategies and create systems that better fit human needs in collaborative tasks. It also helps make the data more organized and useful for analysis or designing weight-sensitive tools.
For weight classification, clustering lets us sort objects based on their weight without needing labels beforehand. We use both unsupervised and supervised methods for classifying weights.

Unsupervised learning attempts to uncover hidden patterns in data without predefined labels. This approach is useful when labeled data is unavailable, making it particularly relevant for exploratory analysis. However, its effectiveness in classification tasks depends on how well natural clusters align with the target categories.

Supervised learning, in contrast, utilizes labeled data to train models that can accurately classify new inputs. By leveraging predefined weight classes, supervised learning methods can achieve high accuracy in weight classification, outperforming unsupervised approaches when sufficient training data is available.

\subsubsection{Unsupervised Learning: K-Means Clustering}

We applied K-Means clustering to classify object weights without labeled data, using both full-arm and hand-only motion data. Initially, our goal was to identify two clusters, corresponding to light and heavy object classes. Afterwards, we extended it to three clusters with an additional medium weight class.
We tested the accuracy of clustering with different weight classes and the results are summarized in Tables \ref{tab:unsupervised_results1}, \ref{tab:unsupervised_results2}.
Using the 3D dominant hand positions as input features from handovers in YCB dataset (excluding objects requiring extra care), K-means achieved at best a moderate accuracy of  81.11\% for a weight classes: Low weight $<$ 1.0 kg and Heavy weight $>$ 2.0 kg.
For other cases of finer classification like Low $<$ 0.1 kg and Heavy $>$ 0.95 kg, the K-Means clustering method achieved an average accuracy around 50\% i.e. close to random guessing, showing that it couldn't effectively classify handovers based on different object weights. 
We further tried the K-Means clustering using the full dominant arm, i.e., the 3D positions of the three body segments of the dominant arm carrying the object: Hand, Lower forearm, and Upper forearm (Table \ref{tab:skeleton1}). Still, the average clustering accuracy was around 50\%. Thus, we can say that it's difficult via unsupervised learning to draw clear boundaries between light, medium, and heavy weighted objects.
\begin{table}[h]
    \centering
    \setlength\abovecaptionskip{-0.07\baselineskip}

    \caption{Unsupervised Learning (K-Means) Accuracy for 2 class Weight Classification}
    \label{tab:unsupervised_results1}
    \begin{tabular}{|c|c|c|}
        \hline
        Weight Classes (Range in kg) & Full Arm & Only Hand \\
        \hline
        2 classes: (0,1.0), (2.0-inf) & 78.31\% & \textbf{81.11}\% \\
        2 classes: (0-0.1), (0.95-inf) & 52.85\% & 55.08\% \\
        \hline
    \end{tabular}
\end{table}
\begin{table}[h]
    \centering
    \setlength\abovecaptionskip{-0.07\baselineskip}

    \caption{Unsupervised Learning (K-Means) Accuracy for 3 class Weight Classification}
    \label{tab:unsupervised_results2}
    \begin{tabular}{|c|c|c|}
        \hline
        Weight Classes (Range in kg) & Full Arm & Only Hand \\
        \hline
        3 classes: (0-1.0), [1.0-2.0], (2.0-inf) & \textbf{67.30}\% & 63.20\% \\
        3 classes: (0-0.1), [0.1-0.95], (0.95-inf) & 50.51\% & 47.31\% \\
        \hline
    \end{tabular}
\end{table}

Thus, K-Means clustering resulted in near-random classification performance for two-class and especially three-class division. The accuracy remained close to 50\% for most cases, indicating that weight classification using unsupervised clustering was ineffective.

\subsubsection{Supervised Learning: KNN, SVM, and Random Forest}
In the supervised learning approach, we utilized labeled weights from handovers within our datasets to train models capable of accurately classifying weight categories. We implemented K-Nearest Neighbors (KNN), Support Vector Machines (SVM), and Random Forest classifiers. 
Each model underwent training and evaluation using stratified cross-validation, ensuring a minimum of 6 splits and 2 repeats to evaluate their performance and reliability. This process also required defining specific weight ranges to create distinct weight classes.
Starting with 2 class classification, the results for both full-arm and hand-only cases are presented in Tables \ref{tab:supervised_full_arm}. Supervised learning significantly outperformed unsupervised clustering, achieving over 90\% accuracy in two-class divisions when using Random Forest. KNN and SVM also performed well, particularly for full-arm data.

\begin{table}[h]
    \centering
    \setlength\abovecaptionskip{-0.07\baselineskip}
    \caption{Supervised Learning Accuracy for 2 weight classes}
    \label{tab:supervised_full_arm}
    \begin{tabular}{|c|c|c|c|}
        \hline
        Weight Classes (Range in kg) & KNN & SVM & Random Forest \\
        \hline
        Full Arm: (0-1.0), [1.00-inf) & 96.89\% & 93.49\% & 97.09\% \\
        Hand: (0-1.0), [1.00-inf) & 97.02\% & 82.33\% & \textbf{97.11}\% \\
        \hline
        Full Arm : (0-0.1), (0.95-inf) & 70.46\% & 70.64\% & 81.25\% \\
        Hand: (0-0.1), (0.95-inf) & 64.59\% & 62.40\% & 72.30\% \\
        \hline
    \end{tabular}
\end{table}

\begin{table}[h]
    \centering
    \setlength\abovecaptionskip{-0.07\baselineskip}
    
    \caption{Supervised Learning Accuracy for 3 weight classes}
    \label{tab:supervised_hand}
     \resizebox{0.98\columnwidth}{!}{
    \begin{tabular}{|c|c|c|c|}
        \hline
        Weight Classes (Range in kg) & KNN & SVM & Random Forest \\
        \hline
        Full Arm : (0-0.1), [0.1-0.95], (0.95-inf) & 57.23\% & 58.01\% & 65.03\% \\
        Hand: (0-0.1), [0.1-0.95], (0.95-inf) & 55.68\% & 40.44\% & 60.49\% \\
         \hline
        Full Arm : (0-1.0), [1.0-2.0], (2.0-inf) & 81.26\% & 64.84\% & \textbf{82.19}\% \\
        Hand: (0-1.0), [1.0-2.0], (2.0-inf) & 80.48\% & 54.03\% & 81.58\% \\
        \hline
    \end{tabular}
    }
\end{table}
Further, we attempted supervised clustering into three weight categories.
However, when applying supervised classification into three weight categories using labels from the previous section, (Low$<$0.099 kg, 0.100 kg$=<$Moderate$=<$0.950 kg, High$>$0.951 kg), we find a maximum accuracy of 65.03\% with full arm and slightly lower accuracy of 60.49\% with just hand positions. 
To get improved accuracy in three weight category classification, we tested different weight ranges as well. We report the results for the weight range (Low$<$0.999 kg, 1.0$=<$Moderate$=<$2.0 kg, High$>$2.0 kg) with which we observe the highest maximum accuracy of 82.19\%.

Overall, for three-class division, accuracy dropped notably, suggesting that finer weight classification remains a challenge. Using only hand data generally resulted in lower accuracy, reinforcing the importance of observing full-arm motion for weight classification. 

\section{Carefulness and Weight}
Prior research has shown that human givers exhibited careful motion in handovers with specific objects and this carefulness can be detected in real-time. We extend this research by studying the compounded effect of weight on carefulness in human motion. For the following analysis, we focused on using the full arm to train classifiers.

\subsection{Classification Accuracy for Four Labels}
Table \ref{tab:four_labels} shows how well different classifiers distinguish between the four labels (cup-careful, cup-notcareful, pitcher-careful, pitcher-notcareful). It is to be noted that the cup is a light weighted object and the pitcher is a heavy weighted object. The Random Forest classifier performs best with an accuracy of 80.5\%, meaning it handles complex patterns well. SVM follows with an accuracy of 68.9\%, using its margin-based decision method. KNN does reasonably well, reaching 64.8\%.

\begin{table}[h]
\centering
\setlength\abovecaptionskip{-0.15\baselineskip}

\caption{Classification Accuracy for Four Labels}

\begin{tabular}{lccc}
\toprule
Classifier & KNN & SVM & Random Forest \\
\midrule
Accuracy  & 64.8\% & 68.9\% & 80.5\% \\
\bottomrule
\end{tabular}
\label{tab:four_labels}
\end{table}

\subsection{Weight-Based Classification}
Table \ref{tab:weight_division} presents the accuracy when objects are split into light and heavy categories. The classification results improve significantly, with Random Forest reaching 90.9\%. This suggests that weight has a strong influence on motion characteristics, making classification easier. SVM also benefits, achieving 85.0\%, while KNN improves to 83.6\%, showing that similar-weight objects share more motion patterns.

\begin{table}[h]
\centering
\setlength\abovecaptionskip{-0.15\baselineskip}
\caption{Classification Accuracy for Weight-Based Division}
\begin{tabular}{lccc}
\toprule
Classifier & KNN & SVM & Random Forest \\
\midrule
Accuracy  & 83.6\% & 85.0\% & 90.9\% \\
\bottomrule
\end{tabular}
\label{tab:weight_division}
\end{table}

\subsection{Carefulness-Based Classification}
Table \ref{tab:carefulness_division} shows classification accuracy when handover motions are grouped by carefulness. While accuracy improves, the effect is less strong compared to weight-based division. Random Forest performs best at 87.1\%, followed by SVM at 76.7\%. KNN achieves 73.8\%, indicating that carefulness does impact motion but not as strongly as weight does.

\begin{table}[h]
\centering
\setlength\abovecaptionskip{-0.15\baselineskip}

\caption{Classification Accuracy for Carefulness-Based Division}
\begin{tabular}{lccc}
\toprule
Classifier & KNN & SVM & Random Forest \\
\midrule
Accuracy  & 73.8 \% & 76.7\% & 87.1\% \\
\bottomrule
\end{tabular}
\label{tab:carefulness_division}
\end{table}

\subsection{Low weight Careful Motion Classification}
Table \ref{tab:cup_classification} shows how well classifiers separate careful and non-careful weighted cup handover motions. Random Forest achieves an excellent accuracy of 94.0\%, suggesting that motion differences for lightweight objects are clear. SVM also performs well at 88.6\%, while KNN reaches 77.7\%, still providing reliable classification.

\begin{table}[h]
\centering
\setlength\abovecaptionskip{-0.15\baselineskip}
\caption{Classification Accuracy for Cup}
\begin{tabular}{lccc}
\toprule
Classifier & KNN & SVM & Random Forest \\
\midrule
Accuracy  & 77.7\% & 88.6\% & 94.0\% \\
\bottomrule
\end{tabular}
\label{tab:cup_classification}
\end{table}

\subsection{High weight Careful Motion Classification}
Table \ref{tab:pitcher_classification} summarizes the accuracy for careful and non-careful heavy weighted pitcher handover motions. The results are lower than for the light weighted cup, likely due to greater motion variability in heavier objects. Random Forest performs best at 79.7\%, followed by SVM at 75.0\% and KNN at 67.5\%, showing moderate classification ability.

\begin{table}[h]
\centering
\setlength\abovecaptionskip{-0.15\baselineskip}
\caption{Classification Accuracy for Pitcher}
\begin{tabular}{lccc}
\toprule
Classifier & KNN & SVM & Random Forest \\
\midrule
Accuracy  & 67.5\% & 75.0\% & 79.7\% \\
\bottomrule
\end{tabular}
\label{tab:pitcher_classification}
\end{table}

\subsection{Conclusion for carefulness and weight}
Dividing by weight greatly improves classification accuracy, showing that object mass plays a key role in motion characteristics. Carefulness-based division also improves accuracy but to a lesser extent, meaning careful and non-careful actions share some similarities. 
Particularly careful handover motion with lighter objects like cups have clearer motion differences compared to heavier ones like pitchers, making them easier to classify. This suggests that as object weight increases, the distinction between careful and non-careful motions becomes less pronounced, making carefulness-based classification more challenging in human handovers.

\section{Conclusion and Future Work}
In this work, we introduced the YCB-Handovers dataset, which provides a standardized benchmark for studying human-robot handovers. This dataset includes a diverse set of objects with well-defined properties, making it useful for reproducible experiments and improving handover strategies. Using this dataset, we analyzed how object characteristics, particularly object weight, influence human-robot handovers. Our results show that object weight impacts both the human motion in handovers as well as the carefulness in human handovers.
Heavier objects tend to result in slower, more deliberate receiving motions, indicating an adaptation in human behavior based on perceived difficulty and risk.

For future work, we plan to extend the dataset by including more participants and testing in different environments to better capture real-world variations. We also aim to explore how robots can adjust their handover strategies based on object properties and human motion. Finally, testing these approaches in real-world applications, such as industrial and assistive robotics, will help improve the practicality and reliability of robotic handovers.
\bibliography{biblio}
\bibliographystyle{IEEEtran}

\vspace{2pt}

\end{document}